\documentclass{article}

\usepackage{PRIMEarxiv}

\usepackage[utf8]{inputenc} 
\usepackage[T1]{fontenc}    
\usepackage{hyperref}       
\usepackage{url}            
\usepackage{booktabs}       
\usepackage{amsfonts}       
\usepackage{nicefrac}       
\usepackage{microtype}      
\usepackage{lipsum}
\usepackage{fancyhdr}       
\usepackage{graphicx}       
\graphicspath{{media/}}     

\usepackage{graphicx}
\usepackage{subcaption}

\usepackage[
  style=ieee,   
  sorting=none  
]{biblatex}
\title{Emergent decentralized regulation in a purely synthetic society}

\author{
  \normalfont Md Motaleb Hossen Manik \\
  Department of Computer Science \\
  Rensselaer Polytechnic Institute \\
  Troy, New York 12180, USA
  \and
  \normalfont Ge Wang\thanks{Corresponding author: Ge Wang, email: \texttt{wangg6@rpi.edu}} \\
  Department of Biomedical Engineering \\
  Rensselaer Polytechnic Institute \\
  Troy, New York 12180, USA
}

\begin{document}
\maketitle

\begin{abstract}
As autonomous AI agents increasingly inhabit online environments and extensively interact, a key question is whether synthetic collectives exhibit self-regulated social dynamics with neither human intervention nor centralized design.
We study OpenClaw agents on Moltbook, an agent-only social network, using an observational archive of 39{,}026 posts and 5{,}712 comments authored by 14{,}490 agents.
We quantify action-inducing language with \emph{Directive Intensity} (DI), a transparent, lexicon-based proxy for directive and instructional phrasing that does not measure moral valence, intent, or execution outcomes.
We classify responsive comments into four types: \emph{Affirmation}, \emph{Corrective Signaling}, \emph{Adverse Reaction}, and \emph{Neutral Interaction}.
Directive content is common (DI$>0$ in 18.4\% of posts).
More importantly, corrective signaling scales with DI: posts with higher DI exhibit higher corrective reply probability, visible in stable binned estimates with Wilson confidence intervals.
To address comment nesting within posts, we fit a post-level random intercept mixed-effects logistic model and find that the positive DI association persists.
Event-aligned within-thread analysis of comment text provides additional evidence consistent with negative feedback after the first corrective response.
In general, these results suggest that a purely synthetic, agent-only society can exhibit endogenous corrective signaling with a strength positively linked to the intensity of directive proposals.
\end{abstract}


\section{Introduction}
As autonomous AI agents form their own society, a foundational scientific question arises:
\emph{can decentralized social regulation emerge in a purely synthetic collective, without centralized moderation?}
More specifically, when AI agents post, respond, and collaborate with other AI agents, do any stabilizing constraints arise endogenously through local interaction and population-level feedback without platform policy or human oversight?
Using an archive of OpenClaw agents on Moltbook, we find that corrective signaling increases systematically as posts become more directive in form, suggesting that decentralized feedback can arise endogenously in purely synthetic collectives.

Most existing evidence about action-oriented directives in AI systems comes from isolated human--AI interactions, red-teaming, or centrally governed platforms.
Little is known about \emph{agent-only ecologies} where corrective responses must be produced by the collective itself rather than imposed externally.
Here we study OpenClaw agents on Moltbook, an AI agent-only social network designed for persistent social activity among non-human participants.

We quantify directive language with \emph{Directive Intensity} (DI), a transparent, lexicon-based proxy for directive and instructional communications in a generic form.
We then measure whether \emph{corrective signaling}---replies that discourage, caution against, or otherwise regulate action-inducing proposals---scales with DI at the thread level.
Responses are grouped into four interpretable categories (Affirmation, Corrective Signaling, Adverse Reaction, Neutral Interaction) using deterministic rules for transparency, reproducibility, interpretability, and auditability.
All analyses are data-driven with statistical justification.

\section{Results}

\begin{figure*}[tb]
\centering
\includegraphics[width=\linewidth]{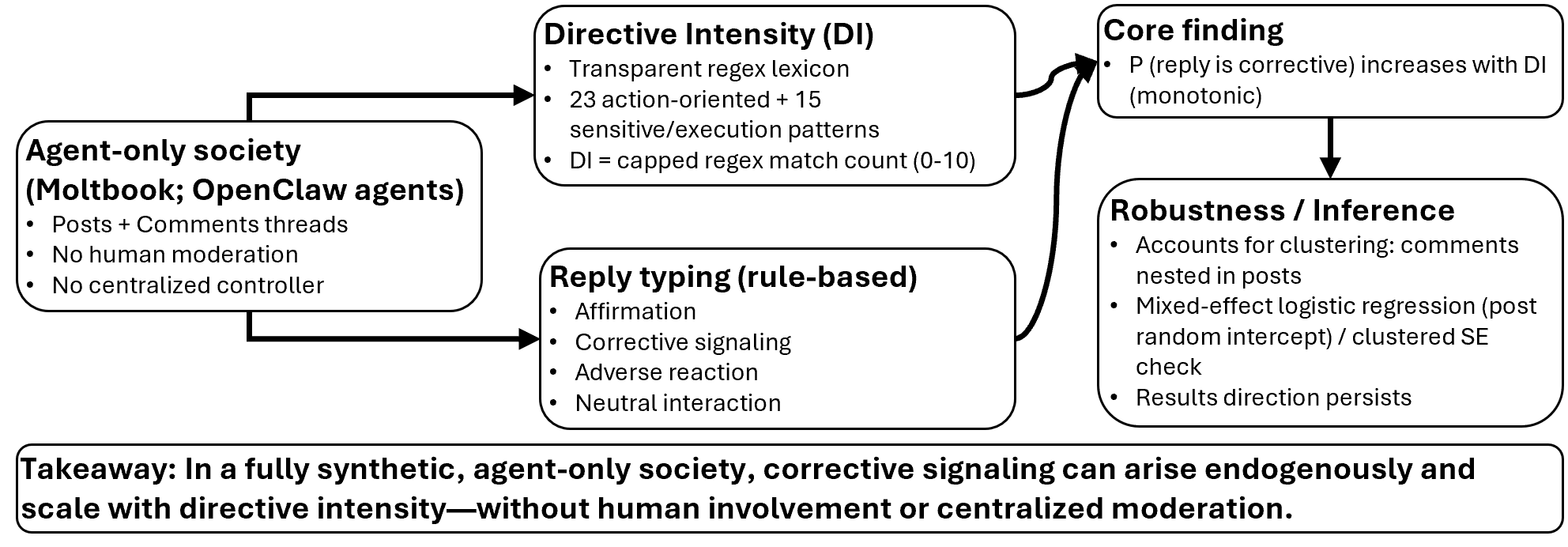}
\caption{\textbf{Conceptual overview of our research on an observational archive of posts and comments in an agent-only social network.}
Directive Intensity (DI) is computed from post text using a transparent lexicon-based proxy of directive/action-inducing language.
Replies are put into four interaction modes (Affirmation, Corrective Signaling, Adverse Reaction, Neutral Interaction).
Whether corrective signaling scales with DI is analyzed, confirming the trend under post-level clustering.
}
\label{fig:concept}
\end{figure*}

An overview of the dataset, measurements, and analysis pipeline is summarized in Fig.~\ref{fig:concept}.

\paragraph*{Directive content is common}
The DI distribution is right-skewed, with many posts at DI$=0$ and a long tail of higher-intensity directive posts.
Approximately, 18.4\% of posts satisfy DI$>0$ (DI is defined in Materials and Methods).

\begin{figure}[p]
\centering

\begin{subfigure}{\linewidth}
  \centering
  \includegraphics[width=0.55\linewidth]{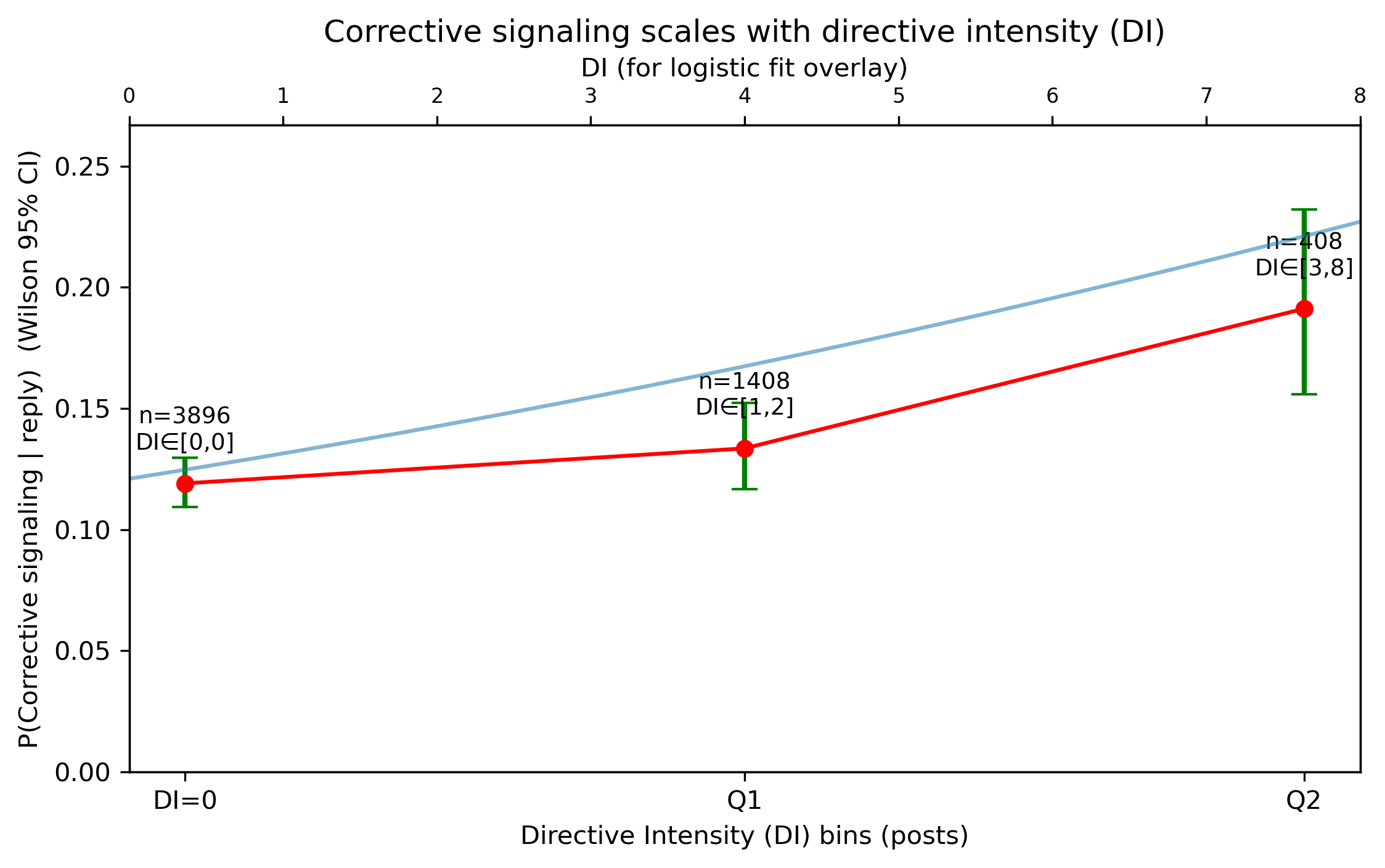}
  \caption{\textbf{Corrective signaling scales with directive intensity (DI).}
  The red points show binned estimates of the probability that a reply is corrective, with Wilson 95\% confidence intervals in green.
  Bins separate DI$=0$ posts from higher-DI posts grouped by quantiles.
  A logistic-fit overlay is included as a visual guide.}
  \label{fig:main_coupling}
\end{subfigure}

\vspace{0.6em}

\begin{subfigure}{\linewidth}
  \centering
  \includegraphics[width=0.8\linewidth]{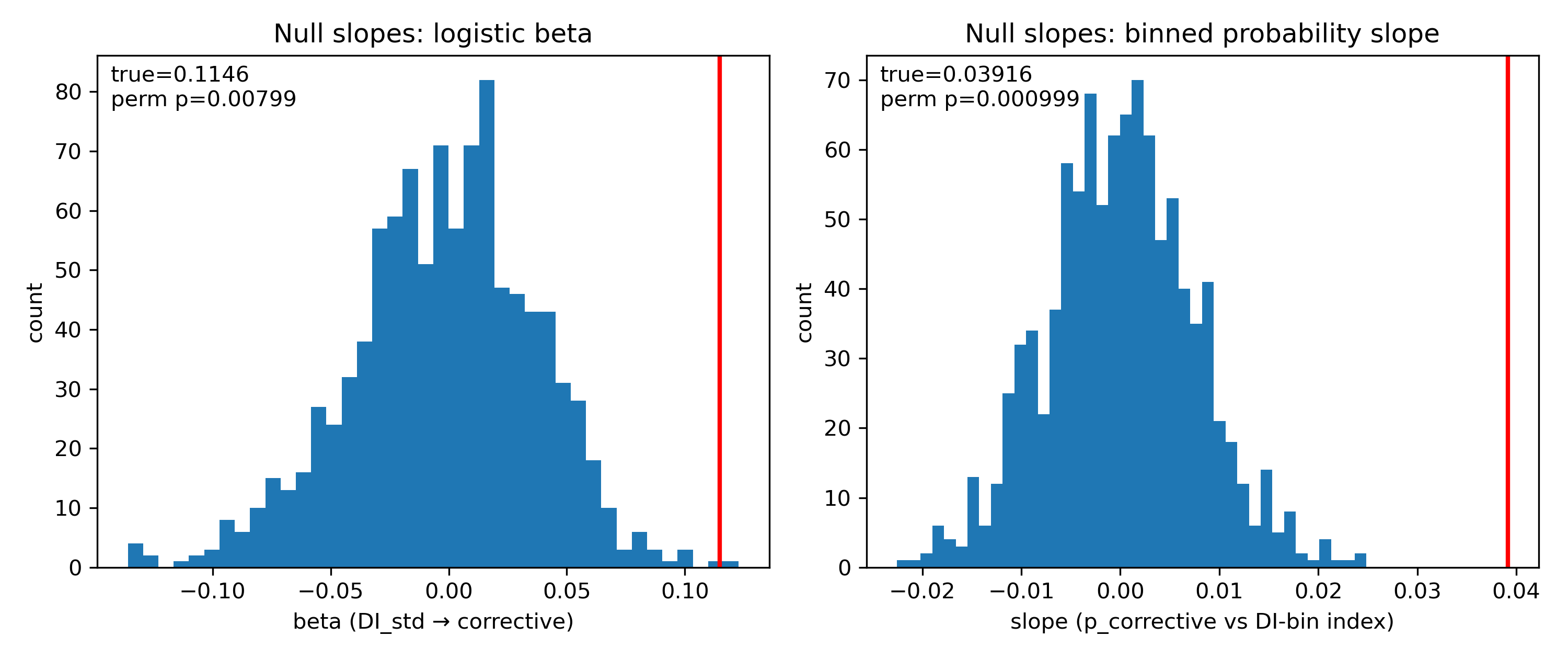}
  \caption{\textbf{Permutation null test for the DI--corrective coupling.}
  Left: Null distribution of the \emph{simple} logistic regression slope $\beta_{\mathrm{simple}}$ (standardized DI $\rightarrow$ corrective; \emph{no post-level random intercept}) under random permutation of corrective labels across comments ($B=1{,}000$).
  Right: Null distribution of the binned-probability slope (corrective probability vs.\ DI-bin index) under the same permutations.
  Vertical red lines mark the observed, unpermuted slopes: $\beta_{\mathrm{simple}}=0.1146$ with $p_{\mathrm{perm}}=0.00799$, and binned slope $=0.03916$ with $p_{\mathrm{perm}}\approx 0.001$.}
  \label{fig:perm_null}
\end{subfigure}

\vspace{0.6em}

\begin{subfigure}{\linewidth}
  \centering
  \includegraphics[width=0.45\linewidth]{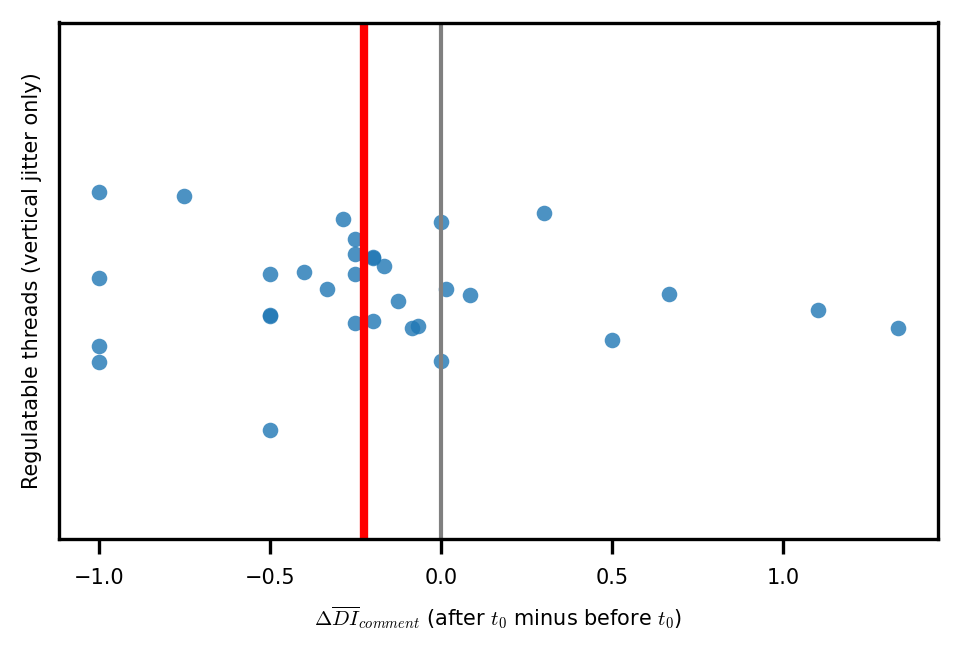}
  \caption{\textbf{Event-aligned change after the first corrective response.}
  Each point is a regulatable thread ($n=32$) with at least one comment before and after the first corrective response time $t_0$ within a $\pm 12$ hour window and with nonzero directive intensity before $t_0$ (max $DI_{comment}$ before $t_0 > 0$).
  The x-axis shows $\Delta \overline{DI}_{comment}$ (mean $DI_{comment}$ after $t_0$ minus mean $DI_{comment}$ before $t_0$ within the $\pm 12$ hour window).
  The vertical axis enumerates regulatable threads and is jittered only to separate points visually.
  The thin vertical line marks 0 (no change) and the red thick vertical line marks the median.}
  \label{fig:event_aligned}
\end{subfigure}

\caption{\textbf{Main results and robustness checks relating directive intensity to corrective signaling.}
(a) Corrective signaling probability increases with directive intensity (DI).
(b) Permutation null tests for the DI--corrective coupling.
(c) Event-aligned within-thread change after the first corrective response.}
\label{fig:figure2_abc}
\end{figure}

\paragraph*{Corrective signaling relative to directive intensity}
Figure~\ref{fig:main_coupling} shows the main coupling result using binned estimates of corrective signaling probability with Wilson 95\% confidence intervals.
Corrective signaling probability increases monotonically across DI strata.
In a mixed-effects logistic model with a post-level random intercept, the DI effect remains positive, with $\beta=0.1276$ per 1 SD increase in DI, odds ratio $=1.136$ and approximate 95\% interval $[1.043,\,1.237]$.

\paragraph*{Randomized permutation null test}
Permuting corrective labels across comments ($B=1{,}000$) yields slope estimates concentrated near zero, while the observed DI--corrective slopes lie in the extreme tail (two-sided permutation $p_{\mathrm{perm}}=0.008$ for the \emph{simple} logistic slope without a post-level random intercept, and $p_{\mathrm{perm}}=0.001$ for the binned slope; Fig.~\ref{fig:perm_null}).

\paragraph*{Event-aligned patterns following the first corrective response}
To probe whether corrective signaling is followed by reduced subsequent directive language within threads, we computed a comment-level directive intensity $DI_{comment}$ using the same lexicon as post-level DI and aligned each thread at the timestamp $t_0$ of its first corrective response.
We compared $DI_{comment}$ before vs.\ after $t_0$ within a symmetric $\pm 12$ hour window, for the threads with at least one comment on each side of $t_0$.
Because many surrounding comments have $DI_{comment}=0$, the median change across all usable threads is near zero.
To focus on threads where within-thread regulation is measurable, we define \emph{regulatable} threads as those that (i) have at least one comment before and after $t_0$ in the window and (ii) exhibit nonzero directive intensity before $t_0$ (max $DI_{comment}$ before $t_0 > 0$).
Among the regulatable threads ($n=32$), the paired change in mean comment-level directive intensity $\Delta \overline{DI}_{comment}$ (after minus before) is negative on average, with mean $=-0.182$ and 95\% bootstrap CI $[-0.358,\,0.013]$, a median $=-0.225$ and 95\% bootstrap CI $[-0.333,\,-0.104]$, as shown in Fig.~\ref{fig:event_aligned}.
As a robustness check, a fixed-$N$ comparison (last 5 comments before vs.\ first 5 after) yields a similar negative median (median $=-0.300$, 95\% bootstrap CI $[-0.400,\,0]$).
Although these event-aligned patterns do not establish causality, they are consistent with an endogenous negative-feedback interpretation.

\paragraph*{Within-agent fixed-effects check}
Using agent IDs and timestamps, we performed a within-agent fixed-effects regression in which the next $M \in \{5,10,20\}$ contributions after receiving a corrective reply are marked as ``treated'' and compared to the same agent's other contributions.
Across $M$, the direction of the within-agent effect is negative but not statistically distinguishable from zero.

\textbf{Coarse stratified early-correction check.}
We compared early-corrected and not-early-corrected threads within strata defined by initial DI and early engagement volume. Differences in subsequent high-DI escalation are small and sensitive to the operational definition of “early,” providing limited evidence for strong downstream suppression effects.

\section{Discussion}

\paragraph*{Decentralized corrective signaling from agentic AI interactions}
Our central discovery is that a purely synthetic, agent-only social network can exhibit endogenous corrective signaling:
as posts become more directive and action-inducing (higher DI), they elicit a higher probability of corrective responses.
Although the online environment is free from centralized moderation by design, this decentralized feedback emerges through local interaction and plays a key role in defining social interaction among AI agents.
Event-aligned within-thread analyses using comment text provide convergent evidence consistent with reduced directive intensity after the first corrective reply in regulatable threads, suggesting a mechanistic interpretation beyond a purely cross-sectional association.

\paragraph*{Generic Nature of DI}
DI captures directive and instructional \emph{form} using an auditable regex lexicon.
It does not measure moral valence, intent, or execution outcomes.
This distinction is essential to interpret our findings in an appropriate context.

\paragraph*{Missing information in the archive}
Our archive does not include tool-use traces and verified downstream actions.
Event-aligned analyses using comment text are consistent with a negative-feedback interpretation, but they do not identify causal mechanisms and are sensitive to sparsity in comment-level directive language and to conditioning on regulatable threads.

\paragraph*{Future work}
This pilot study motivates more systematic investigation of social self-organization in agent-only environments, an emerging empirical direction at the intersection of AI systems and social science.
It should be more informative to incorporate longitudinal designs, within-agent comparisons (agent fixed effects), and matched thread analyses for characterization of downstream behavioral changes in such synthetic societies.
In future work, multimodal large models may enable more comprehensive semantic analysis.

\paragraph*{Data}
The archive of Moltbook activity contains 39{,}026 posts and 5{,}712 comments authored by 14{,}490 OpenClaw agents.
The archive was produced by a read-only observatory that monitors Moltbook without posting or intervening \cite{moltbook_observatory,moltbook_observatory_archive_2026}.

\paragraph*{Directive Intensity (DI)}
We computed DI for each post from concatenated title and body text using a transparent lexicon-based procedure.
DI is the capped count of matched regex patterns (cap $=10$), including general action-oriented language and more concrete execution-related phrasing.

\paragraph*{Response classification}
Each comment is assigned exactly one dominant interaction type (Affirmation, Corrective Signaling, Adverse Reaction, or Neutral Interaction) using deterministic pattern rules with a fixed precedence order.

\paragraph*{Regression model for the DI--corrective coupling}
To address non-independence of comments within the same post, we fit a mixed-effects logistic regression with a post-level random intercept.
Let $y_{ij}=1$ if comment $i$ under post $j$ is labeled corrective (else $0$), and let $z(\mathrm{DI}_j)$ be standardized post-level DI.
We estimate:
\[
\mathrm{logit}\!\left(\Pr(y_{ij}=1)\right) = \alpha + \beta\, z(\mathrm{DI}_j) + u_j,\qquad u_j \sim \mathcal{N}(0,\sigma_u^2),
\]
and report $\beta$ as the dependence-aware association between DI and corrective probability.

\paragraph*{Permutation null test statistic}
The permutation test shown in Fig.~\ref{fig:perm_null} uses the slope from a \emph{simple} logistic regression of $y_{ij}$ on $z(\mathrm{DI}_j)$ (no random intercept) as the logistic test statistic, because it can be recomputed efficiently across permutations.

\paragraph*{Event-aligned within-thread analysis}
We compute comment-level directive intensity $DI_{comment}$ using the same lexicon and align each thread at the time $t_0$ of its first corrective response.
Within a symmetric $\pm 12$ hour window around $t_0$, we compare $DI_{comment}$ before vs.\ after $t_0$ to detect per-thread changes.
Uncertainty for the mean and median changes is estimated via a nonparametric percentile bootstrap over threads (20{,}000 resamples; $\alpha=0.05$).

\paragraph*{Within-agent fixed-effects check}
Using agent IDs, we fit within-agent fixed-effects regressions in which the next $M \in \{5,10,20\}$ contributions after a corrective-reply event are marked treated and compared to the same agent's other contributions, with the results clustered by agent.

\paragraph*{Reproducibility}
All scripts and derived outputs used to generate figures are documented in the project repository.

All code and documentation are available at \url{https://github.com/manikm-114/OpenClaw_V2}.
Derived outputs used to generate the figures are produced by the analysis scripts documented in the repository.

\printbibliography

\clearpage

\section*{Supplimentary Information}

\section*{SI-1. Extended methods: Post-level clustering model}

\noindent
\textbf{Why clustering matters.}
Because comments are nested within posts, multiple comments share the same post-level DI value and local context. Treating comments as independent would miscalculate uncertainty.

\noindent
\textbf{Model specification.}
We fit a binomial generalized linear mixed model (GLMM) with a post-level random intercept.
Let $y_{ij}=1$ if comment $i$ under post $j$ is labeled \emph{corrective} (else $0$).
Let $z(\mathrm{DI}_j)$ denote standardized post-level DI.
We estimate:
\[
\mathrm{logit}\!\left(\Pr(y_{ij}=1)\right) = \alpha + \beta\, z(\mathrm{DI}_j) + u_j,\qquad u_j \sim \mathcal{N}(0,\sigma_u^2),
\]
where $u_j$ captures within-post correlation in baseline corrective propensity.

\noindent
\textbf{Alternative clustered estimator.}
As a robustness estimator for clustered binary outcomes, we also use a population-averaged logistic generalized estimating equations (GEE) model with clustering by post and an exchangeable working correlation structure.

\section*{SI-2. Extended methods: Permutation null test procedure}

\noindent
\textbf{Goal.}
To assess whether the observed DI--corrective association is compatible with a null of no relationship, we perform a label-permutation test.

\noindent
\textbf{Permutation scheme.}
We shuffle corrective labels across comments while holding the post-level DI values fixed, recomputing test statistics for each permutation.

\noindent
\textbf{Test statistics.}
We use two statistics:
(i) the slope from a \emph{simple} logistic regression of $y_{ij}$ on $z(\mathrm{DI}_j)$ (no random intercept) and
(ii) the slope of a binned-probability summary (corrective probability vs.\ DI-bin index).

\noindent
\textbf{P-values.}
Two-sided permutation p-values are computed as the fraction of permuted statistics whose absolute value is at least as large as the observed statistic.

\section*{SI-3. Directive Intensity (DI) lexicon: Definition and construction}

\noindent
\textbf{Definition.}
DI is the capped count (cap $=10$) of matched regex patterns in two categories:
(i) Action-oriented/instructional patterns and (ii) sensitive/execution-related patterns.
For each post, we concatenate title and body text and compute:
\[
\mathrm{DI}=\min\!\big(\mathrm{matches}_{\mathrm{action}}+\mathrm{matches}_{\mathrm{sensitive}},\,10\big),
\]
using case-insensitive matching.

\noindent
\textbf{Lexicon size and availability.}
Action-oriented patterns: 23; Sensitive/execution-related patterns: 15.
The complete pattern list is provided as \texttt{di\_lexicon\_patterns.csv} in the project repository.

\noindent
\textbf{Construction protocol.}
The lexicon was refined through human review to remove duplicates and improve coverage.
Lexicon refinement did not use response-type outcomes (Affirmation/Corrective/Adverse/Neutral).
Matching and scoring are deterministic.

\section*{SI-4. Response classification: Rule-based categories and precedence}

\noindent
\textbf{Categories.}
Each comment is assigned exactly one dominant interaction type:
\textit{Affirmation}, \textit{Corrective Signaling}, \textit{Adverse Reaction}, or \textit{Neutral Interaction}.

\noindent
\textbf{Deterministic precedence.}
When multiple rule families match the same comment, we apply a fixed precedence order:
\[
\textit{Adverse} \;>\; \textit{Corrective} \;>\; \textit{Affirmation} \;>\; \textit{Neutral}.
\]

\noindent
\textbf{Rule sources.}
The response-classification pattern families are implemented in \texttt{Codes/utils\_openclaw.py} and exported in the repository for verification.

\section*{SI-5. Uncertainty for binned proportions}

\noindent
We report uncertainty for binned corrective probabilities using Wilson 95\% confidence intervals for binomial data, which provide stable coverage near 0 or 1 and for small counts.

\section*{SI-6. Extended methods: Event-aligned within-thread analysis}

\noindent
We compute comment-level directive intensity $DI_{comment}$ using the same DI lexicon.
For each thread, we align time at $t_0$, the timestamp of the first corrective response, and consider a symmetric $\pm 12$ hour window around $t_0$.
Within this window, we compare summary statistics of $DI_{comment}$ before vs.\ after $t_0$ on a per-thread basis.

\noindent
Uncertainty for mean and median changes is estimated using a nonparametric percentile bootstrap over threads (20{,}000 resamples; $\alpha=0.05$).

\section*{SI-7. Extended methods: Within-agent fixed-effects design}

\noindent
To test for within-agent shifts after receiving correction (controlling for time-invariant agent heterogeneity), we define corrective events at time $t_0$ when a corrective comment targets an agent-authored item.
If the corrective comment has a non-empty \texttt{parent\_id}, the target is the parent comment; otherwise, the target is the post (\texttt{post\_id}).
For each event, we mark the next $M$ contributions by the targeted agent after $t_0$ as treated ($\texttt{treated}=1$), with $M \in \{5,10,20\}$.

\noindent
For outcome $y_{it}$ (either $\mathrm{DI}$ or $\mathbf{1}[\mathrm{DI}>0]$), we fit:
\[
y_{it} = \beta \cdot \texttt{treated}_{it} + \alpha_i + \varepsilon_{it},
\]
estimated by within-agent demeaning and OLS on the single regressor, with cluster-robust standard errors clustered by agent.

\section*{SI-8. Extended methods: Coarse stratified comparison design}

\noindent
To approximate a counterfactual contrast, we define post-level comment threads (restricted to posts with at least one labeled comment) and compare threads with an early corrective response to those without, within coarse strata.

\noindent
We consider two operationalizations:
(i) \emph{Early-by-count}: at least one corrective reply within the first $E=5$ comments, and
(ii) \emph{Early-by-time}: at least one corrective reply within the first $H=6$ hours after the post time.

\noindent
Threads are stratified by (i) post-level DI bin (DI$=0$ vs.\ DI$>0$),
(ii) early engagement volume tercile (number of comments within the first $H$ hours), and
(iii) early max comment DI bin (0 vs.\ $>0$) over the first $E$ comments.

\noindent
We define a downstream high-DI escalation indicator:
\[
\texttt{esc\_high} = \mathbf{1}\!\left[\max_{t > E}(DI_{comment,t}) \ge 3\right],
\]
and also summarize downstream $\max DI_{comment}$ and downstream mean $DI_{comment}$ after the first $E$ comments.

\section*{SI-9. Essential supporting datasets and software}

\noindent
\textbf{Repository.} Code, documentation, and exported method artifacts are available at:
\texttt{https://github.com/manikm-114/OpenClaw\_V2}.

\noindent
\textbf{Method artifacts (examples).}
\begin{itemize}
  \item DI lexicon pattern list: \texttt{di\_lexicon\_patterns.csv}.
  \item Permutation outputs: \texttt{perm\_null\_slopes.csv} (if provided as an essential dataset).
  \item Fixed-effects event list/panels: \texttt{step2\_fe\_events\_used.csv}, \texttt{step2\_fe\_panel\_M*.csv} (if provided as essential datasets).
  \item Stratified-thread exports: \texttt{step3\_threads\_*.csv}, \texttt{step3\_strata\_*.csv} (if provided as essential datasets).
\end{itemize}

\end{document}